\documentclass[letterpaper, 10 pt, conference]{ieeeconf}  

\IEEEoverridecommandlockouts                              

\usepackage{amssymb}
\usepackage{amsmath}
\let\labelindent\relax
\usepackage{enumitem}
\usepackage{enumitem}
\usepackage{graphicx}
\usepackage[dvipsnames,table]{xcolor}
\usepackage{tikz}
\usetikzlibrary{arrows.meta,bending,positioning,calc,math,backgrounds,fit}
\usepackage{booktabs}
\usepackage{algorithm}
\usepackage{algorithmicx}
\usepackage{algpseudocode}
\usepackage{multirow}
\usepackage[separate-uncertainty=true]{siunitx}
\usepackage[font=footnotesize]{caption}
\usepackage[font=footnotesize]{subcaption}

\usepackage[hidelinks]{hyperref}
\definecolor{hmred}{HTML}{E89B8B}
\definecolor{hmyel}{HTML}{F2E2A0}
\definecolor{hmgrn}{HTML}{A0CD91}
\newcommand{\cl}[1]{\cellcolor{hmyel!#1!hmred}} 
\newcommand{\ch}[1]{\cellcolor{hmgrn!#1!hmyel}} 
\usepackage{enumitem}

\title{\LARGE \bf
Not All Layers Need Tuning: Diagnosing and Directing Adaptation in Vision-Language-Action Models
}

\author{
    \textbf{Shahram Najam Syed}\textsuperscript{1},
    \textbf{Arthur Jakobsson}\textsuperscript{1},
    \textbf{Prayuj Sachdev}\textsuperscript{2},  
    \textbf{Jeffrey Ichnowski}\textsuperscript{1} \\
    \textsuperscript{1}Robotics Institute, Carnegie Mellon University, Pittsburgh, USA \\
    \textsuperscript{2}Department of Electrical and Computer Engineering, Carnegie Mellon University, Pittsburgh, USA
}

\makeatletter
\g@addto@macro\normalsize{%
  \setlength\abovedisplayskip{5pt plus 2pt minus 2pt}%
  \setlength\belowdisplayskip{5pt plus 2pt minus 2pt}%
  \setlength\abovedisplayshortskip{3pt}%
  \setlength\belowdisplayshortskip{3pt}%
}
\makeatother

\begin{document}

\maketitle
\thispagestyle{empty}
\pagestyle{empty}

\begin{abstract}%
Fine-tuning a Vision-Language-Action (VLA) model for a new deployment environment is expensive, yet most methods apply uniform-capacity adapters to every network region as if every region requires equal adjustment. This paper tests that assumption on five architecturally diverse VLAs (OpenVLA-OFT, $\pi_0$, SmolVLA, DTP, Octo; 93M--7B parameters). Measuring per-region adaptation cost as normalized parameter displacement under region-isolated fine-tuning reveals an adaptation spectrum in which appearance shifts concentrate cost in the vision encoder, instruction shifts in the language backbone, and novel-object shifts in the vision encoder together with the action head, consistently across all five architectures in our experiments. To exploit this structure, we introduce a pipeline that observes, diagnoses, allocates, and adapts. From ten unlabeled target observations and without fine-tuning, the diagnostic estimates per-region cost by combining reference-free gradient and Monte Carlo (MC) Dropout signals with a Centered Kernel Alignment (CKA) score against a cached source reference; the allocator converts the estimates into variable-rank Low-Rank Adaptation (LoRA) adapters under a parameter budget and freezes well-calibrated regions; and standard LoRA fine-tuning adapts the policy with the resulting adapters. In our experiments, the diagnostic ranks regions within each deployment at a median Spearman of 0.91, and the allocation matches or exceeds uniform LoRA at every budget we tested on LIBERO and CALVIN. On a physical xArm-7, the pipeline reaches 21/30 instruction-shift successes against 22/30 for full fine-tuning with 0.04\% of its trainable parameters, and on five held-out scenes evaluated without retraining it leads every baseline, with 11--23 successes of 30 rollouts against 8--18 for the strongest parameter-efficient baseline at equal or larger budgets and 2--11 for full fine-tuning. These results suggest that adaptation cost in VLAs is structured enough to measure before fine-tuning begins.
\end{abstract}

\section{Introduction}
\label{sec:introduction}

Vision-Language-Action (VLA) models~\cite{Kim2024OpenVLA, OctoModelTeam2024} degrade under deployment-time distribution shifts in environment appearance, object geometry, or instruction phrasing, and recovering reliable behavior requires adaptation~\cite{Ferchau2026VLALoRA}.
Every deployment site presents its own shift, so adaptation repeats across sites, and each instance operates in a low-data regime, typically tens of demonstrations collected on the robot.
Full fine-tuning updates billions of parameters from only tens of demonstrations, a gap of more than eight orders of magnitude, and in our physical experiments it yields lower held-out-scene success than selective adaptation under a shared fixed-epoch training protocol.
The dominant strategy applies parameter-efficient fine-tuning (PEFT) methods such as Low-Rank Adaptation (LoRA)~\cite{Hu2022LoRA} uniformly across all layers at a fixed global rank, which implicitly assumes that every network region requires equal adjustment.
A VLA, however, comprises functionally distinct regions, namely a vision encoder, a language backbone, cross-modal projectors, and an action head, and a perturbation confined to one input modality need not implicate every region.
Biological sensorimotor control exhibits the same structure, as the nervous system is thought to maintain separable internal models and to update only those whose predictions fail~\cite{Wolpert1998MOSAIC}.
When a shift implicates only a subset of regions, uniform allocation overprovisions the unimplicated regions and underprovisions the implicated ones.

\begin{figure}[!tp]
    \centering
    \setlength{\abovecaptionskip}{4pt}
    \setlength{\belowcaptionskip}{-15pt}
    \includegraphics[width=\columnwidth]{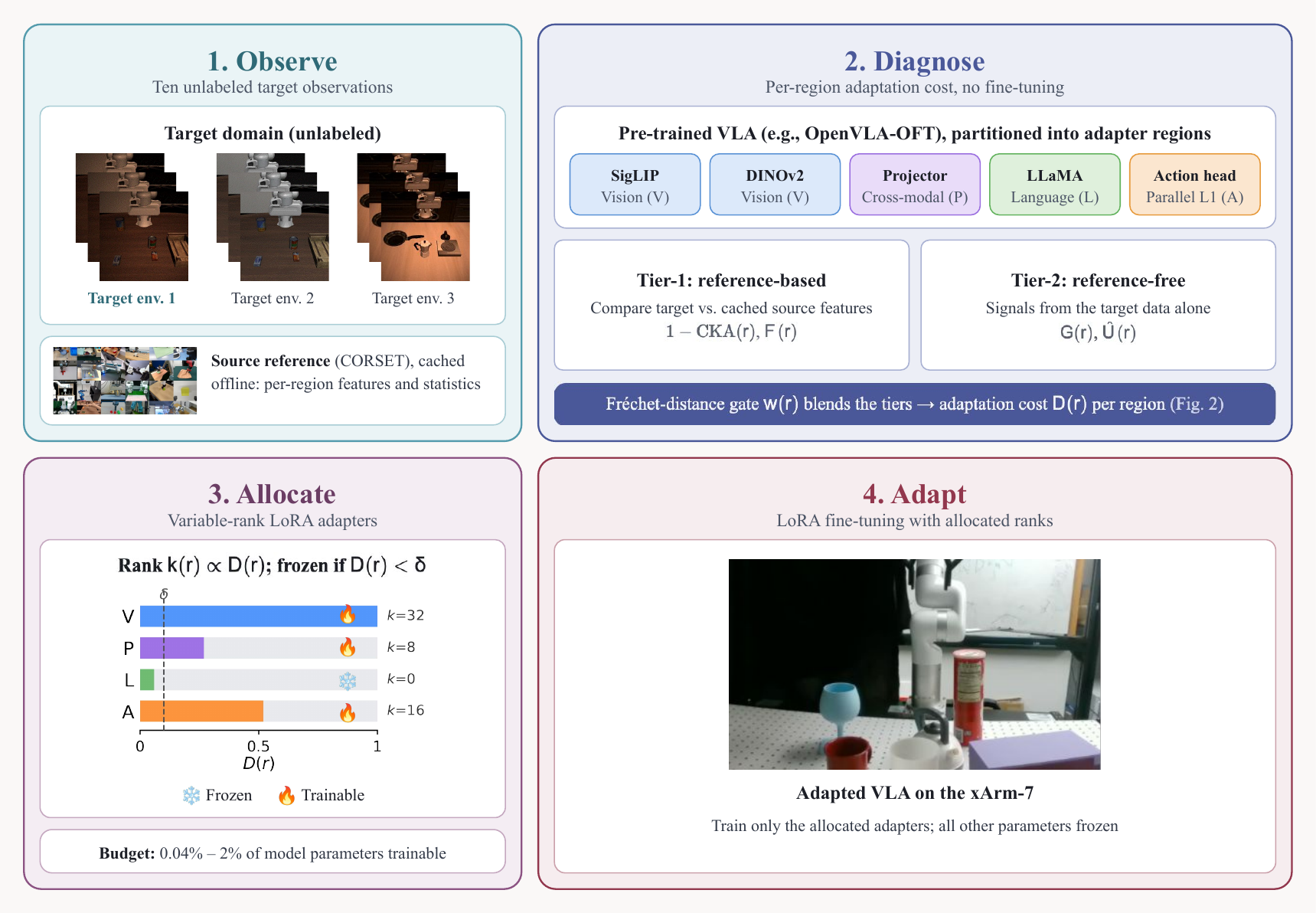}
    \caption{\textbf{Observe, diagnose, allocate, adapt.} From a few unlabeled target observations, the pipeline estimates per-region adaptation cost using Tier-1 reference-based signals, Tier-2 reference-free signals, and a Fr\'echet-distance gate to blend them. A budget-aware allocator assigns variable-rank LoRA adapters and freezes well-calibrated regions, and the adapted policy reaches 21/30 instruction-shift successes on the xArm-7 at a small fraction of the trainable parameters.}
    \label{fig:system-overview}
\end{figure}

Selective adaptation has precedent in vision models~\cite{Lee2023Surgical, He2023SPT, Zhang2023AdaLoRA}, and concurrent work allocates non-uniform LoRA capacity during VLA fine-tuning~\cite{Kim2026LoRASP} or traces VLA representations after fine-tuning~\cite{Shi2026VLATrace}.
To the best of our knowledge, this is the first work to measure, before fine-tuning and from a small set of unlabeled target observations, how adaptation demand distributes across the functional regions of a VLA, and to convert that measurement into a capacity allocation.

This paper first characterizes that distribution empirically across five architecturally distinct VLAs under five controlled deployment shifts.
The resulting \emph{adaptation spectrum} concentrates adaptation cost in a shift-specific region, with the dominant region consistent across all five architectures in our experiments.
We then introduce a pipeline with four stages, which \emph{observes}, \emph{diagnoses}, \emph{allocates}, and \emph{adapts} (Fig.~\ref{fig:system-overview}, \S\ref{sec:method}).
We evaluate the pipeline on the LIBERO~\cite{Liu2023LIBERO} and CALVIN~\cite{Mees2022CALVIN} benchmarks and on a physical UFactory xArm-7 manipulator.
The diagnostic ranks regions within each deployment at a median Spearman correlation of $0.91$, and the allocation matches or exceeds uniform LoRA at every parameter budget we tested on all five architectures.
On the xArm-7, the pipeline reaches 21/30 instruction-shift successes against 22/30 for full fine-tuning with 0.04\,\% of its trainable parameters, and on five held-out scenes evaluated without retraining it leads every baseline, with 11--23 successes of 30 against 8--18 for the strongest PEFT baseline at equal or larger budgets and 2--11 for full fine-tuning.
This paper makes three contributions.
\begin{enumerate}[itemsep=2pt, topsep=3pt, partopsep=0pt]
    \item An empirical characterization of how adaptation cost distributes across functional regions under five deployment shifts, with the dominant region per shift type consistent across five VLA architectures (93M--7B parameters) and three action-decoding paradigms.

    \item An algorithm that estimates per-region adaptation cost from ten unlabeled target observations without fine-tuning and converts the estimates into variable-rank LoRA adapters matching uniform rank-32 success rates at a reduced parameter count in our experiments.

    \item Experiments across five architectures, five seeds, and six physical deployment shifts that characterize where the allocation helps and include one negative result in which it does not.
\end{enumerate}
\section{Related Work}
\label{sec:related-work}

\textbf{VLA adaptation.}
VLA models pair pretrained vision-language backbones with autoregressive~\cite{Brohan2023RT1, Kim2024OpenVLA, Kim2025OpenVLAOFT}, diffusion~\cite{OctoModelTeam2024, Hou2024DTP}, or flow-matching~\cite{Black2024Pi0, Shukor2025SmolVLA} action decoders, and current practice adapts them with one global LoRA rank across all layers~\cite{Kim2024OpenVLA, Kim2025OpenVLAOFT}.
Building on this parameterization, the pipeline keeps LoRA as the adapter, whose low-rank form suits the low intrinsic dimensionality of fine-tuning~\cite{Aghajanyan2021Intrinsic}, and replaces the single global rank with a per-region allocation.
A study of LoRA for $\pi_0$~\cite{Ferchau2026VLALoRA} reports that uniform allocation suffices at rank~32; our results (\S\ref{sec:exp-allocation}) agree at that budget and diverge below it, where the adaptive advantage concentrates.
LoRA also forgets less than full fine-tuning~\cite{Biderman2024LoRALearnsLess}, a regularization we observe on hardware (\S\ref{sec:exp-physical}).

\textbf{Non-uniform rank allocation.}
AdaLoRA~\cite{Zhang2023AdaLoRA} prunes singular values, ARD-LoRA~\cite{ARDLoRA2025} learns per-head rank schedules, and LoRA-SP~\cite{Kim2026LoRASP} gates singular value decomposition (SVD) update directions, all during training.
In contrast to these methods, which require a full optimization run to find the allocation, the pipeline fixes ranks before training from unlabeled observations, and we compare against AdaLoRA and LoRA-SP at matched budgets (\S\ref{sec:exp-allocation}).
Surgical fine-tuning~\cite{Lee2023Surgical} showed in vision models that the shift type determines which layers benefit from tuning, and SPT~\cite{He2023SPT} and RSRA~\cite{Liu2026RSRA} also assign adaptation capacity before fine-tuning, from gradient or representation sensitivity.
The contribution here is to characterize region-wise adaptation demand across heterogeneous VLAs and to estimate it from ten unlabeled target observations, combining source-referenced representational shift with reference-free target sensitivity, and to evaluate the resulting allocation on physical hardware.

\textbf{Representational and uncertainty measures.}
The diagnostic scores source-target alignment with Centered Kernel Alignment (CKA)~\cite{Kornblith2019CKA}, gates its tiers with the Fr\'{e}chet distance~\cite{Dowson1982Frechet}, and estimates uncertainty with Monte Carlo (MC) Dropout~\cite{Gal2016MCDropout}.
Its Fisher weighting is stored compactly with Kronecker-factored approximate curvature (KFAC)~\cite{Martens2015KFAC}.
The contribution lies in combining these measures and applying them before adaptation.
VLA-Trace~\cite{Shi2026VLATrace} also applies CKA to VLA representations; in contrast to its post-hoc analysis of fine-tuned models, the diagnostic runs before adaptation and drives a capacity allocation.
\section{Problem Statement}
\label{sec:problem}

Consider a pretrained VLA, $\pi_\theta$, that maps an observation $o \in \mathcal{O}$ (RGB images and, in some architectures, proprioceptive state) and a language instruction $\ell \in \mathcal{L}$ to an action or action chunk $a \in \mathcal{A} \subseteq \mathbb{R}^d$.
The parameters $\theta \in \mathbb{R}^P$, with $P$ the total parameter count, partition into $R$ disjoint, exhaustive \emph{network regions} $\theta = (\theta^{(1)}, \dots, \theta^{(R)})$ with $\theta^{(r)} \in \mathbb{R}^{P_r}$ and $\sum_{r=1}^{R} P_r = P$, each a functional component (e.g., vision encoder, language backbone, cross-modal projector, action head) whose boundaries the architecture fixes.

\paragraph{Inputs and goal}
Let $\mathcal{D}_\text{tgt}$ denote a target deployment distribution that differs from the pretraining source distribution $\mathcal{D}_\text{src}$ in visual appearance, spatial layout, object geometry, language phrasing, or a combination of these.
Given $N$ unlabeled target observations $\mathcal{T}_\text{diag} = \{(o_j, \ell_j)\}_{j=1}^{N}$, a demonstration set $\mathcal{T} = \{(o_i, \ell_i, a_i)\}_{i=1}^{M}$ from $\mathcal{D}_\text{tgt}$ with $N < M$, and a trainable-parameter budget $P_\text{budget}$, the goal is to adapt $\pi_\theta$ into a policy $\pi_{\theta'}$ that maximizes task success rate on $\mathcal{D}_\text{tgt}$ while training at most $P_\text{budget}$ parameters.
The output is a per-region LoRA rank assignment $k : \{1, \dots, R\} \to \mathbb{Z}_{\geq 0}$, with $k(r) = 0$ a frozen region, that satisfies
\begin{equation}
\label{eq:budget-constraint}
    \sum_{r=1}^{R} k(r)\, c_1(r) \leq P_\text{budget},
\end{equation}
where $c_1(r)$ is the parameter cost of one unit of rank in region $r$ (\S\ref{sec:method-allocate}), together with the adapters trained under that assignment.

\paragraph{Adaptation cost}
To evaluate whether an assignment places capacity where it is needed, we define a per-region reference.
Let $\theta^{(r)\star}_{\mathcal{T}}$ denote the parameters that region-isolated fine-tuning produces, i.e., fine-tuning region $r$ on $\mathcal{T}$ to convergence while holding all other regions fixed.
The \emph{per-region adaptation cost} is the normalized parameter displacement this adaptation induces,
\begin{equation}
\label{eq:adaptation-cost}
    \rho(r) = \frac{
        \bigl\lVert \theta^{(r)\star}_{\mathcal{T}} - \theta^{(r)} \bigr\rVert_2
    }{
        \bigl\lVert \theta^{(r)} \bigr\rVert_2
    } \;\in\; \mathbb{R}_{\geq 0}.
\end{equation}
A small $\rho(r)$ indicates little parameter displacement under this region-isolated protocol, so $\rho(r)$ is an operational proxy for adaptation demand rather than a measure of causal necessity.
Our experiments treat $\{\rho(r)\}_{r=1}^{R}$ as the reference signal for evaluating the diagnostic.

\paragraph{Assumptions}
Beyond these inputs, we assume access to (i)~the pretrained weights with forward and backward passes and (ii)~a compact source reference constructed offline and available at deployment (\S\ref{sec:method-diagnose}), comprising a small coreset of source observations and their per-region embedding statistics.
We assume neither knowledge of the shift type at adaptation time nor deployment-time access to source training data beyond that reference.
\section{Method}
\label{sec:method}

The pipeline runs the four stages of Fig.~\ref{fig:system-overview} in order, each consuming the output of the last.
The \emph{observe} stage collects $\mathcal{T}_\text{diag}$ at the deployment site.
The \emph{diagnose} stage maps $(\pi_\theta, \mathcal{T}_\text{diag})$ to per-region scores $D(r) \in (0, 1]$ through a comparison tier against a cached source reference and a reference-free tier, blended by a per-region Fr\'{e}chet gate (\S\ref{sec:method-diagnose}); the scores must agree with $\rho(r)$ in rank order within a deployment and in calibrated magnitude across deployments, since the allocator consumes both.
The \emph{allocate} stage maps the scores to the rank assignment $k(r)$ of Eq.~\ref{eq:budget-constraint}, proportional to $D(r)$ up to per-region cost and zero below a freezing threshold (\S\ref{sec:method-allocate}).
The \emph{adapt} stage trains LoRA adapters of the assigned ranks on $\mathcal{T}$ (\S\ref{sec:method-adapt}).
Each stage runs once per deployment, and only the adapt stage updates parameters.

\subsection{Network Region Partitioning}
\label{sec:method-regions}

Adaptation cost is measured, and adapters are attached, per region, so the partition determines both what the diagnostic can resolve and what the allocator can control.
We partition each VLA along its native module boundaries into the functional regions of Table~\ref{tab:region-partitions} rather than imposing a fixed schema, for three reasons.
Native boundaries coincide with changes in representational role, from modality-specific encoding through fusion to action decoding, so a shift confined to one modality maps onto few regions; they carry architecture-defined parameter counts, which fixes $c_1(r)$ without tuning; and they exist for every architecture, which makes the spectrum comparable across the five models without a learned or hand-designed partition per model.
Finer partitions raise diagnostic cost and make the per-region Fisher and gradient statistics noisier, and coarser ones merge regions with different roles, so the module level is the natural granularity.

\begin{table}[tbp]
    \centering
    \setlength{\abovecaptionskip}{0pt}
    \setlength{\belowcaptionskip}{2pt}
    \caption{Region partitions with approximate parameter counts. AR, Diff., and FM denote autoregressive, diffusion, and flow-matching decoding.}
    \label{tab:region-partitions}
    \scriptsize
    \renewcommand{\arraystretch}{0.92}
    \setlength{\tabcolsep}{3pt}
    \resizebox{\columnwidth}{!}{%
    \begin{tabular}{@{}lcl@{}}
        \toprule
        \textbf{Model (Params)} & \textbf{Dec.} & \textbf{Regions} \\
        \midrule
        OpenVLA-OFT (7B) & AR (par.) & DINOv2, SigLIP, Proj., LLaMA-e, LLaMA-l, Act.\ Head \\
        $\pi_0$ (3.3B)   & FM        & SigLIP, Gemma, Act.\ Expert \\
        SmolVLA (450M)   & FM        & SigLIP, SmolLM2, Proj., Act.\ Expert \\
        DTP (350M)       & Diff.     & DINOv2, CLIP, Q-Former, Diff.\ Trans. \\
        Octo (93M)       & Diff.     & Vis.\ Tok., T5 Enc., Transformer, Diff.\ Head \\
        \bottomrule
    \end{tabular}}
\end{table}

\subsection{Stages 1--2: Observe and the Two-Tier Diagnostic}
\label{sec:method-diagnose}

\begin{figure}[!tp]
    \centering
    \setlength{\abovecaptionskip}{4pt}
    \setlength{\belowcaptionskip}{-15pt}
    \includegraphics[width=0.9\columnwidth]{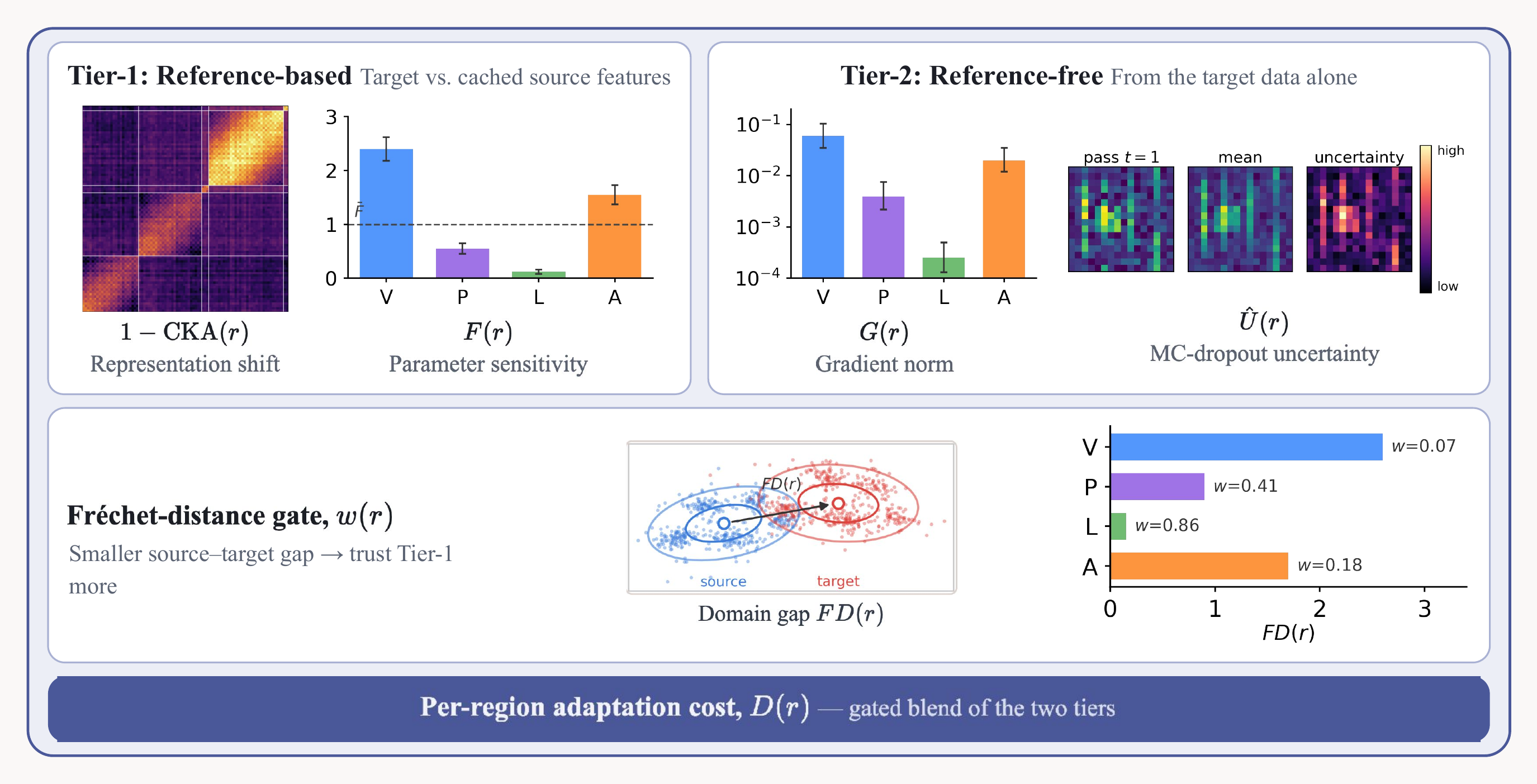}
    \caption{\textbf{The two-tier diagnostic.} Tier~1 scores representational shift ($1 - \mathrm{CKA}(r)$) against the cached source reference, weighted by Fisher sensitivity $\overline{F}(r)$ normalized to a global mean of one (dashed line); Tier~2 profiles the surrogate gradient norm $G(r)$ and MC Dropout uncertainty $\hat{U}(r)$ from target observations alone; the gate $w(r) = \exp(-\gamma \cdot \mathrm{FD}(r))$ trusts Tier~1 where the source--target gap is small, and the blend yields $D(r)$. V, P, L, and A denote vision encoder, projector, language backbone, and action head.}
    \label{fig:diagnostic}
\end{figure}

From the $N$ unlabeled observations $\mathcal{T}_\text{diag}$, the two-tier diagnostic estimates per-region cost with a small number of forward passes and one backward pass through the \emph{frozen} VLA, routing each region to the strongest available signal for its domain gap.

\paragraph{Cached source reference}
To provide a source-domain reference without deployment-time access to source training data, a cache is built once per architecture offline and is available at deployment.
We select a K-means coreset~\cite{HarPeled2007Coresets} of $K$ observations over per-region output embeddings of an $N_\text{pool}$-observation candidate pool and store, per region, the activation mean and covariance, a buffer of $B$ feature vectors for CKA, and an empirical Fisher under the model's native objective as KFAC factors~\cite{Martens2015KFAC} truncated to their top $k_F$ eigencomponents.
Coreset selection outperforms random sampling because the comparison signals below are sensitive to coverage of the source manifold (\S\ref{sec:exp-ablations}).
With the values of \S\ref{sec:exp-setup}, the reference occupies 200--250~MB for the 7B model and 30--50~MB for sub-1B models.

\paragraph{Fr\'{e}chet gate}
To decide which tier each region trusts, the gate computes the Fr\'{e}chet distance~\cite{Dowson1982Frechet} $\mathrm{FD}(r)$ between the cached source Gaussian and the empirical target distribution at the region's output layer, normalized by the layer's feature dimension so that scales are comparable across regions of different width.
A soft weight $w(r) = \exp(-\gamma \cdot \mathrm{FD}(r))$ with temperature $\gamma$ then sets how much each region trusts the comparison tier.
Every observation contributes an activation vector at each token position, so the estimates draw on thousands of feature vectors rather than $N$ samples, and we apply shrinkage to the target covariance before the matrix square root.

\paragraph{Tier 1 (comparison)}
When $w(r)$ is high, the diagnostic scores representational shift with CKA~\cite{Kornblith2019CKA} between the cached source activations and the target activations.
Because the two sets are unpaired and of different sizes, we evaluate a covariance-alignment analogue of linear CKA, the normalized Frobenius inner product between their centered second-moment matrices.
This statistic is well defined without pairing, invariant to isotropic scaling and common rotations, and blind to mean shifts, which are exactly the component the Fr\'{e}chet gate measures, so the two are complementary by construction.
To fund a shifted representation only when it also affects the output, the tier weights the CKA term by parameter sensitivity,
\begin{equation}
\label{eq:tier1-score}
    D_\text{cmp}(r) = \bigl(1 - \mathrm{CKA}(r)\bigr) \cdot \overline{F}(r),
\end{equation}
where $\overline{F}(r)$ is the mean diagonal Fisher over the region's LoRA-eligible parameters, normalized by the global mean.
A reliability factor $r_\text{rel}(N) = \min(1, N/N_0)$ downweights Tier~1 below a reference count $N_0$, with $\mathrm{CKA}(r) = 0$ at $N < 2$ where the estimator is undefined.
The gate keeps Tier~1 active only when the domain gap is small, which is the regime in which the source-derived Fisher remains a valid sensitivity estimate.

\paragraph{Tier 2 (reference-free)}
When $w(r)$ is low, the diagnostic profiles the size-normalized gradient norm of a label-free surrogate loss matched to each model's native objective,
\begin{equation}
\label{eq:gradient-score}
    G(r) = \frac{\bigl\lVert \nabla_{\theta^{(r)}} \mathcal{L}_\text{surr} \bigr\rVert_2}{\sqrt{\dim(\theta^{(r)})}},
\end{equation}
with pseudo-labels detached through stop-gradient.
Diffusion and flow-matching objectives retain stochastic targets, so their gradients remain nonzero under pseudo-labeling.
For the deterministic L1 head of OpenVLA-OFT, a loss against the model's own detached prediction would vanish identically, so the surrogate perturbs the detached pseudo-action with Gaussian noise of scale $\sigma$ in normalized action units, which makes the gradient a random-sign projection of the action Jacobian, $-J_\theta^{\top}\operatorname{sign}(\epsilon)$, and its per-region norm a label-free measure of local output sensitivity rather than of any task objective.
The LLaMA backbone uses cross-entropy against its own detached token predictions, and upstream regions receive gradients from these losses through backpropagation.
For Octo and DTP, which train with dropout, MC Dropout~\cite{Gal2016MCDropout} with dropout rate $p$ over $T$ stochastic passes supplies an activation-variance score $U(r)$.
Hats denote division by one global maximum per signal, taken over all regions and source-side calibration scenarios (\S\ref{sec:exp-setup}) and clipped to $1$, so scores lie in $(0, 1]$ and a mild shift can score low in every region.
The tier score is
\begin{equation}
\label{eq:tier2-score}
    D_\text{fall}(r) = \alpha\,\hat{U}(r) + (1 - \alpha)\,\hat{G}(r),
\end{equation}
where $\alpha > 0$ only for dropout-trained models, since $U(r)$ is undefined otherwise.

\paragraph{Combined score}
With the Tier~1 score normalized in the same way to $\hat{D}_\text{cmp}(r) \in (0, 1]$, the diagnostic score blends the tiers,
\begin{equation}
\label{eq:combined-score}
    D(r) = w(r)\, r_\text{rel}(N)\, \hat{D}_\text{cmp}(r)
         + \bigl(1 - w(r)\, r_\text{rel}(N)\bigr)\, D_\text{fall}(r).
\end{equation}
The blend also degrades gracefully, since $N = 1$ forces near-total reliance on Tier~2, near-uniform scores recover uniform allocation, and Tier~1 carries the prediction when the surrogate mismatches the training objective.

\subsection{Stage 3: Adaptive Rank Allocation}
\label{sec:method-allocate}

We attach LoRA adapters to the attention matrices of Transformer regions and the linear layers of multi-layer perceptron (MLP) regions~\cite{Hu2022LoRA}.
For region $r$ with adapted matrices $\mathcal{I}_r$ of input/output dimensions $\{(d^{\text{in}}_i, d^{\text{out}}_i)\}$, the parameter cost per unit rank is $c_1(r) = \sum_{i \in \mathcal{I}_r} (d^{\text{in}}_i + d^{\text{out}}_i)$.
Under a fixed budget $P_\text{budget}$, the allocator assigns
\begin{equation}
\label{eq:rank-allocation}
    k(r) =
    \begin{cases}
        0 & \text{if } D(r) < \delta, \\
        \max\!\Bigl(1,\;
            \Bigl\lfloor
                \frac{D(r)}{\sum_{r'} D(r')}
                \cdot \frac{P_\text{budget}}{c_1(r)}
            \Bigr\rfloor
        \Bigr) & \text{otherwise,}
    \end{cases}
\end{equation}
where $\delta$ is a freezing threshold, so the allocator freezes a region only when its score falls below a fraction $\delta$ of the calibrated maximum, and the $\max(1, \cdot)$ clause preserves gradient flow through every funded adapter.
Because $c_1(r)$ spans orders of magnitude across regions, Eq.~\ref{eq:rank-allocation} makes the allocated \emph{capacity} $k(r) \cdot c_1(r)$, rather than the raw rank, proportional to the diagnostic score.
The simulation experiments fix $P_\text{budget}$ externally for matched-budget comparison.
In deployment, the allocator instead assigns each region above $\delta$ a rank $k(r) = \max(1, \lfloor \beta \cdot D(r) \rfloor)$ through a global rank-per-score gain $\beta$.
Because $D(r) \leq 1$, the gain caps the per-region rank at $\beta$, so with $\beta = 32$ the allocator reaches parity with uniform rank-$32$ LoRA when every region requires adaptation, and the total budget emerges as the sum of per-region costs and tracks shift severity.
When the rank floors of Eq.~\ref{eq:rank-allocation} exceed a fixed budget, the allocator freezes funded regions in ascending score order until the floors fit and assigns any capacity left over after rounding to the highest-scoring region, so it meets reported budgets within the stated tolerance.

\subsection{Stage 4: Fine-Tuning}
\label{sec:method-adapt}

Fine-tuning runs standard LoRA training on the demonstration set $\mathcal{T}$ with the variable rank per region as the sole modification.
Frozen regions carry no adapter and incur no additional memory or computation.
\section{Experiments}
\label{sec:experiments}

We ask whether adaptation cost varies systematically across regions and shifts (\S\ref{sec:exp-spectrum}), whether the diagnostic predicts it (\S\ref{sec:exp-diagnostic}), whether the allocation converts prediction into task success (\S\ref{sec:exp-allocation}--\S\ref{sec:exp-ablations}), and how it fares on hardware (\S\ref{sec:exp-physical}).

\definecolor{ckaVisA}{HTML}{3E6AA8}
\definecolor{ckaVisB}{HTML}{7FA8D9}
\definecolor{ckaBridge}{HTML}{D9AE3E}
\definecolor{ckaLang}{HTML}{5A9E5A}
\definecolor{ckaAct}{HTML}{C44E52}
\newcommand{\ckasw}[1]{\textcolor{#1}{\rule{4.5pt}{4.5pt}}}
\newcommand{\ckaW}{0.172\textwidth}
\newcommand{\ckahead}[1]{%
    \begin{minipage}[c]{\ckaW}\centering
        \textbf{\footnotesize #1}%
    \end{minipage}}
\newcommand{\ckaarch}[2]{%
    \begin{minipage}[c]{0.07\textwidth}\centering
        \textbf{\scriptsize #1}\\{\tiny (#2)}%
    \end{minipage}}
\newcommand{\ckacell}[1]{%
    \begin{minipage}[c]{\ckaW}\centering
        \includegraphics[width=0.97\linewidth]{images/cka_cells/#1}%
    \end{minipage}}

\begin{figure}[tp]
    \centering
    \setlength{\abovecaptionskip}{4pt}
    \setlength{\belowcaptionskip}{-10pt}
    \setlength{\tabcolsep}{1.4pt}
    \resizebox{\columnwidth}{!}{%
    \begin{minipage}[c]{0.945\textwidth}
        \centering
        \begin{tabular}{cccccc}
             &
            \ckahead{Appearance} &
            \ckahead{Spatial} &
            \ckahead{Novel Object} &
            \ckahead{Instruction} &
            \ckahead{Combined} \\[2pt]
            \ckaarch{OpenVLA-OFT}{7B} &
            \ckacell{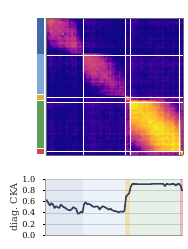} & \ckacell{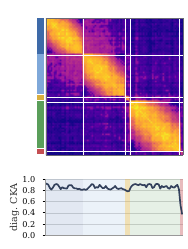} &
            \ckacell{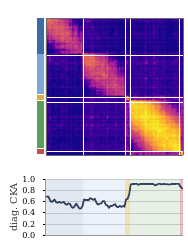} & \ckacell{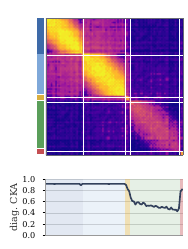} &
            \ckacell{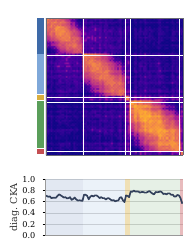} \\
            \ckaarch{$\pi_0$}{3.3B} &
            \ckacell{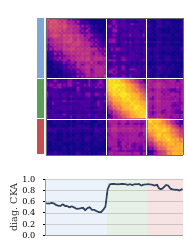} & \ckacell{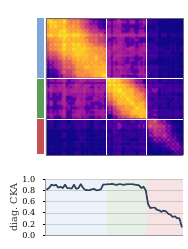} &
            \ckacell{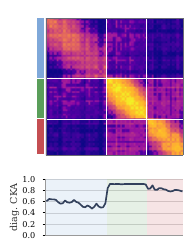} & \ckacell{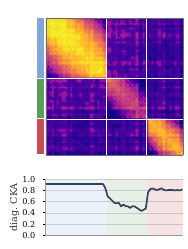} &
            \ckacell{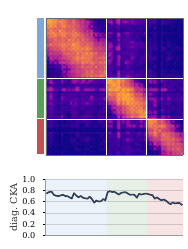} \\
        \end{tabular}\\[3pt]
        {\scriptsize
        \ckasw{ckaVisA}\,\ckasw{ckaVisB}~vision encoders \quad
        \ckasw{ckaBridge}~cross-modal bridge \quad
        \ckasw{ckaLang}~language backbone \quad
        \ckasw{ckaAct}~action head}
    \end{minipage}%
    \begin{minipage}[c]{0.1\textwidth}
        \includegraphics[width=\linewidth]{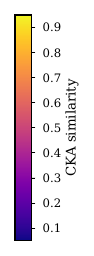}
    \end{minipage}}
    \caption{Source--target representational similarity for OpenVLA-OFT and $\pi_0$ (rows) under the five deployment shifts of Table~\ref{tab:shift-construction} (columns); the remaining three architectures appear in the accompanying video. Each matrix is linear CKA~\cite{Kornblith2019CKA} between source- and target-layer activations on \emph{paired} observations (the same LIBERO scenes rendered under both conditions), well defined across layers of different width. Role-colored bands mark the regions of Table~\ref{tab:region-partitions}. The trace beneath each matrix is the same-layer Tier~1 statistic (\S\ref{sec:method-diagnose}), the unpaired covariance-alignment score against the cached source reference that the diagnostic consumes. In each column the same role darkens across both architectures.}
    \label{fig:cka-table}
\end{figure}

\subsection{Setup}
\label{sec:exp-setup}

\paragraph{Architectures and benchmarks}
We evaluate the five VLA architectures of Table~\ref{tab:region-partitions}, which span three action-decoding paradigms and two orders of magnitude in parameter count.
LIBERO~\cite{Liu2023LIBERO} is the primary benchmark ($40$ manipulation tasks across four suites), and CALVIN~\cite{Mees2022CALVIN} the secondary, which tests transfer to a different task distribution.
Within each benchmark we construct four single-axis shifts and their combined perturbation (Table~\ref{tab:shift-construction}).

\begin{table}[tbp]
    \centering
    \setlength{\abovecaptionskip}{0pt}
    \setlength{\belowcaptionskip}{2pt}
    \caption{The five controlled shifts, each modifying one axis of the observation distribution, or all four in the combined shift, while preserving task semantics.}
    \label{tab:shift-construction}
    \scriptsize
    \renewcommand{\arraystretch}{0.92}
    \setlength{\tabcolsep}{3pt}
    \begin{tabular}{@{}l p{\dimexpr\columnwidth-1.75cm\relax}@{}}
        \toprule
        \textbf{Shift} & \textbf{Construction} \\
        \midrule
        Appearance   & Warm tint, vignette, contrast, desaturation, and Gaussian noise on rendered RGB \\
        Spatial      & Distractor clutter and partial occlusion of the workspace at scene reset \\
        Novel Object & Mesh and geometry substitution with new material, color, and reflectance \\
        Instruction  & LLM rephrasing under semantic equivalence, $10\%$ manually verified \\
        Combined     & All four perturbations applied jointly \\
        \bottomrule
    \end{tabular}
\end{table}

\paragraph{Baselines}
We compare against eleven baselines in three families and refer to the allocation the pipeline produces as \emph{Adaptive}.
\emph{No Adaptation} (the frozen pretrained policy) is the floor.
\emph{Full Fine-Tuning}, \emph{Uniform LoRA} at rank~$32$, and \emph{Oracle Allocation} (the allocator driven by the measured $\rho(r)$ instead of the diagnostic) are nominal upper bounds.
\emph{Uniform LoRA matched}, \emph{DoRA matched}~\cite{Liu2024DoRA}, \emph{$(IA)^3$}~\cite{Liu2022FewShot}, \emph{AdaLoRA}~\cite{Zhang2023AdaLoRA}, \emph{LoRA-SP}~\cite{Kim2026LoRASP}, \emph{Surgical} (the full matched budget on the spectrum's dominant region, an oracle-informed variant of surgical fine-tuning~\cite{Lee2023Surgical}), and \emph{Random Allocation} compete at a matched trainable-parameter budget.

\paragraph{Protocol and statistics}
We collect $M = 20$ demonstrations per task in the target environment and give the diagnostic $N = 10$ unlabeled observations.
All methods train for $50$ epochs with BFloat16 precision, cosine learning-rate decay, gradient clipping, and AdamW, with 8-bit optimizer states and gradient checkpointing for Full Fine-Tuning of the two largest models.
We evaluate every method at its best validation checkpoint, which we select by action-prediction loss on held-out demonstrations, so we score no method in its overfitting tail.
We run $5$ seeds per condition and $50$ rollouts per task and report mean $\pm$ standard error of the mean (SEM).
We compare methods with paired t-tests, paired by task, with Bonferroni correction across architectures, and we report all conditions, including those where the pipeline ties with or trails a baseline.
We set the cache constants to $K = 300$, $N_\text{pool} = 5{,}000$, $B = 200$--$500$ (by architecture), and $k_F = 100$, the diagnostic constants to $N_0 = 10$, $\sigma = 0.05$, $p = 0.1$, $T = 20$, and $\alpha = 0.5$ for dropout-trained models, and we calibrate the gate temperature $\gamma = 1$, the freezing threshold $\delta = 0.1$, and the rank-per-score gain $\beta = 32$ once on source-side scenarios from the LIBERO training split, which also supplies the cached reference; we hold every value fixed across all architectures, shifts, benchmarks, and physical experiments, so no target-domain information enters hyperparameter selection.
We tune learning rates per method family (full fine-tuning, the LoRA family, and $(IA)^3$) on the same scenarios, since a shared learning rate would disadvantage full fine-tuning relative to LoRA.

\subsection{The Adaptation Spectrum}
\label{sec:exp-spectrum}

This experiment measures $\rho(r)$ of Eq.~\ref{eq:adaptation-cost} by exhaustive region-isolated fine-tuning (\S\ref{sec:problem}) for every (architecture, shift, seed) condition on both benchmarks, and provides the reference signal for evaluating the diagnostic.
Region-isolated fine-tuning can inflate $\rho(r)$ for high-capacity regions that compensate for errors arising elsewhere, and displacement depends on the optimization protocol as well as on necessity.
Two checks support $\rho(r)$ as an allocation target despite this caveat.
Oracle Allocation (\S\ref{sec:exp-allocation}), which converts $\rho(r)$ directly into ranks, approaches the full fine-tuning bound, and per-region success recovery on the same runs correlates with $\rho(r)$ at a median per-cell Spearman of $\rho_S = 0.87$.

\begin{figure}[tbp]
    \centering
    \setlength{\abovecaptionskip}{4pt}
    \setlength{\belowcaptionskip}{-10pt}
    \includegraphics[width=\columnwidth]{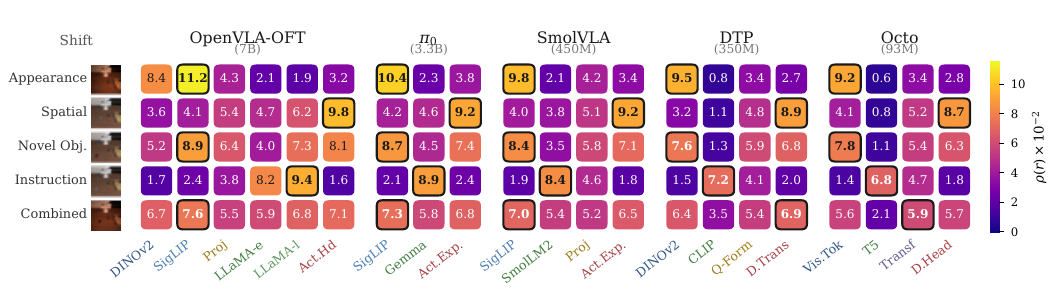}
    \caption{The adaptation spectrum on LIBERO, showing per-region cost $\rho(r) \times 10^{-2}$ (mean over $5$ seeds) with shifts as rows and each architecture's regions as columns. Region labels are role-colored to match Fig.~\ref{fig:cka-table}, and the outlined tile marks the dominant region. CALVIN identifies the same dominant region in all $25$ cells.}
    \label{fig:spectrum-heatmaps}
\end{figure}

Fig.~\ref{fig:spectrum-heatmaps} shows the spectrum on LIBERO.
First, $\rho(r)$ varies by roughly an order of magnitude across regions within a single (architecture, shift) cell, so uniform allocation necessarily overprovisions some regions and underprovisions others.
Second, the dominant region for each shift type is consistent across all five architectures, with appearance shifts loading the vision encoders, instruction shifts the language backbone, and novel-object shifts the vision encoder together with the action head.
The novel-object shift substitutes geometry as well as material and color (Table~\ref{tab:shift-construction}), so recovery requires re-binding perception to new grasp geometry, and the physical shift shows the same signature (\S\ref{sec:exp-physical}).
Third, CALVIN identifies the same dominant region as LIBERO in all $25$ (architecture, shift) cells, with a mean per-cell Spearman of $\rho_S = 0.93$ between the two cost profiles, which suggests the spectrum reflects the architecture--shift interaction rather than a benchmark artifact.
The spectrum is a descriptive finding for the architectures and benchmarks studied here, and we do not claim it extends beyond them.

\subsection{Diagnostic Accuracy}
\label{sec:exp-diagnostic}

\begin{figure}[!tp]
    \setlength{\abovecaptionskip}{4pt}
    \setlength{\belowcaptionskip}{-8pt}
    \centering
    \includegraphics[width=0.348\columnwidth]{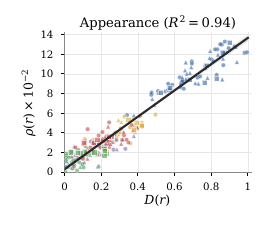}\hfill
    \includegraphics[width=0.300\columnwidth]{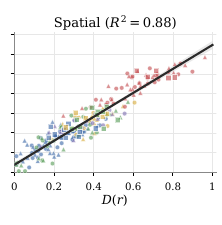}\hfill
    \includegraphics[width=0.300\columnwidth]{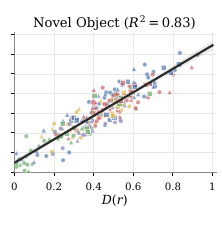}\\[3pt]
    \includegraphics[width=0.300\columnwidth]{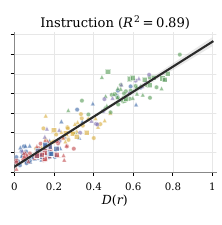}\hspace{0.024\columnwidth}%
    \includegraphics[width=0.300\columnwidth]{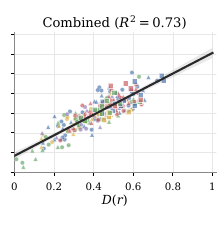}\\[3pt]
    \includegraphics[width=\columnwidth]{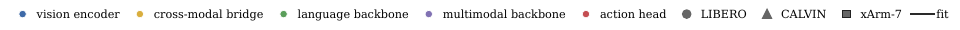}
    \caption{Predicted score $D(r)$ against measured cost $\rho(r)$ by shift type. Each point is one region of one architecture, benchmark, and seed, colored by region role, and squares overlay the physical xArm-7 measurements with OpenVLA-OFT. The combined panel spans a narrower cost range, which lowers its per-panel $R^2$ but not the pooled $R^2 = 0.89$ (Table~\ref{tab:diagnostic-aggregate}).}
    \label{fig:diagnostic-scatter-facets}
\end{figure}

This experiment evaluates whether the diagnostic predicts $\rho(r)$ from $N = 10$ unlabeled observations without fine-tuning.
We report two complementary statistics.
The pooled $R^2$ of predicted scores $\{D(r)\}$ against measured costs $\{\rho(r)\}$ across all $1{,}050$ seed-level points ($210$ per shift) summarizes the fit in Fig.~\ref{fig:diagnostic-scatter-facets}, but it inherits between-region variance, since vision encoders score high and language backbones low across most deployments, and therefore overstates within-deployment discrimination.
The median per-cell Spearman $\rho_S$ over each (architecture, shift, benchmark, seed) cell's regions measures the within-deployment ordering the allocator consumes, though a correlation over $3$--$6$ regions is individually coarse.
We read the two jointly.
The CKA matrices of Fig.~\ref{fig:cka-table} show the raw Tier~1 signal, with each shift darkening the blocks whose regions require adaptation.

\begin{table}[tbp]
    \centering
    \caption{Diagnostic accuracy across the five architectures with $N = 10$ unlabeled target observations and no fine-tuning. $R^2$ is the pooled fit of predicted scores against measured costs over all conditions, and $\rho_S$ is the median per-cell Spearman correlation. Brackets give the range of each statistic across the five architectures, computed per architecture. Bold marks the best result per column.}
    \label{tab:diagnostic-aggregate}
    \scriptsize
    \setlength{\tabcolsep}{2.8pt}
    \begin{tabular}{lcc}
        \toprule
        \textbf{Diagnostic Variant} & \textbf{Spearman $\rho_S$} & \textbf{$R^2$} \\
        \midrule
        \multicolumn{3}{l}{\textit{Na{\"i}ve baselines}} \\
        Random scores                       & 0.02 [0.01, 0.03] & 0.01 [0.00, 0.01] \\
        Frobenius norm      & 0.33 [0.29, 0.37] & 0.18 [0.15, 0.21] \\
        Cosine dist.\ of means & 0.40 [0.36, 0.44] & 0.22 [0.19, 0.25] \\
        \midrule
        \multicolumn{3}{l}{\textit{Tier ablations}} \\
        Tier~1 (CKA $+$ Fisher) & 0.77 [0.71, 0.83] & 0.59 [0.51, 0.66] \\
        Tier~2 (Grad.\ $+$ MC Drop.) & 0.82 [0.77, 0.88] & 0.68 [0.61, 0.74] \\
        Uniform blend (no gate) & 0.87 [0.82, 0.92] & 0.78 [0.70, 0.85] \\
        \midrule
        \textbf{Two-tier (proposed)}        & \textbf{0.91 [0.86, 0.96]} & \textbf{0.89 [0.80, 0.93]} \\
        \bottomrule
    \end{tabular}
\end{table}

Table~\ref{tab:diagnostic-aggregate} and Fig.~\ref{fig:diagnostic-scatter-facets} support three conclusions.
First, na{\"i}ve distributional measures (Frobenius norm, cosine distance) fall far short of either tier.
Second, neither tier alone matches the gated blend, and the ordering holds within every architecture (bracketed ranges in Table~\ref{tab:diagnostic-aggregate}).
Third, the uniform $0.5$ blend trails the gated blend on every architecture.

\subsection{Adaptive Allocation Performance}
\label{sec:exp-allocation}

This experiment asks whether the allocation converts prediction accuracy into task success at reduced parameter cost.

\begin{table}[tbp]
    \centering
    \setlength{\abovecaptionskip}{0pt}
    \setlength{\belowcaptionskip}{2pt}
    \caption{Task success rate ($\%$) on the LIBERO combined shift, mean $\pm$ SEM over $5$ seeds and $50$ rollouts per task across $40$ tasks. Column headers give each architecture's matched budget, about half of Uniform LoRA at rank~$32$. Matched-budget methods meet it within $\pm 5\%$, $(IA)^3$ runs at its native $0.05$M--$0.5$M, and Full Fine-Tuning updates all parameters. Bold marks the best matched-budget result per column. $\dagger$ and $\ddagger$ mark results significantly better than Uniform LoRA matched and AdaLoRA at $p<0.05$ (paired t-tests across tasks, Bonferroni-corrected).}
    \label{tab:allocation-consolidated}
    \renewcommand{\arraystretch}{0.92}
    \resizebox{\columnwidth}{!}{%
    \begin{tabular}{@{}lccccc@{}}
        \toprule
        \textbf{Method} & \textbf{OpenVLA-OFT} (16M) & $\boldsymbol{\pi_0}$ (12M) & \textbf{SmolVLA} (3.4M) & \textbf{DTP} (2.6M) & \textbf{Octo} (1.4M) \\
        \midrule
        No Adaptation           & 23.4$\pm$3.6 & 25.2$\pm$3.4 & 19.8$\pm$2.8 & 21.4$\pm$1.9 & 18.6$\pm$4.9 \\
        Full Fine-Tuning        & 74.6$\pm$1.2 & 76.8$\pm$1.3 & 67.2$\pm$1.1 & 62.7$\pm$1.2 & 58.4$\pm$1.0 \\
        Uniform LoRA $r{=}32$   & 71.2$\pm$2.3 & 73.4$\pm$2.4 & 64.1$\pm$2.1 & 60.3$\pm$2.7 & 56.1$\pm$1.5 \\
        Oracle Allocation       & 73.8$\pm$1.4 & 75.1$\pm$0.8 & 65.7$\pm$1.0 & 61.4$\pm$0.9 & 57.8$\pm$0.9 \\
        \midrule
        Uniform LoRA matched    & 62.4$\pm$4.3 & 64.8$\pm$5.6 & 51.4$\pm$6.1 & 49.8$\pm$3.7 & 47.2$\pm$4.7 \\
        DoRA matched~\cite{Liu2024DoRA} & 64.7$\pm$2.1 & 67.4$\pm$3.7 & 53.8$\pm$4.0 & 52.1$\pm$5.4 & 49.5$\pm$5.7 \\
        $(IA)^3$~\cite{Liu2022FewShot}  & 51.2$\pm$8.1 & 53.4$\pm$6.0 & 42.6$\pm$6.4 & 41.5$\pm$5.0 & 38.4$\pm$10.8 \\
        Random Allocation       & 42.8$\pm$7.7 & 44.2$\pm$5.9 & 35.8$\pm$8.1 & 33.2$\pm$11.5 & 31.6$\pm$10.1 \\
        Surgical (dominant region) & 56.1$\pm$3.6 & 58.4$\pm$4.2 & 45.2$\pm$3.1 & 43.8$\pm$4.8 & 41.5$\pm$3.9 \\
        AdaLoRA~\cite{Zhang2023AdaLoRA} & 67.3$\pm$3.8 & 69.5$\pm$3.5 & 57.9$\pm$2.7 & 55.7$\pm$1.6 & 52.7$\pm$2.9 \\
        LoRA-SP~\cite{Kim2026LoRASP} & 67.2$\pm$2.4 & 71.8$\pm$1.8 & 59.9$\pm$4.9 & 59.4$\pm$1.9 & 46.3$\pm$2.1 \\
        \midrule
        \textbf{Adaptive} & \textbf{73.4$\pm$1.3}$^{\dagger\ddagger}$ & \textbf{76.3$\pm$1.2}$^{\dagger\ddagger}$ & \textbf{64.6$\pm$1.1}$^{\dagger\ddagger}$ & \textbf{61.5$\pm$1.1}$^{\dagger\ddagger}$ & \textbf{57.9$\pm$1.5}$^{\dagger\ddagger}$ \\
        \bottomrule
    \end{tabular}}
\end{table}

Table~\ref{tab:allocation-consolidated} reports success rates on the LIBERO combined-shift environment for all five architectures, and Fig.~\ref{fig:pareto-facets} traces the parameter--success frontier under each of the five shift types.
Four observations follow.
First, at the matched budget, Adaptive outperforms every matched-budget baseline on every architecture in our experiments, with significance against both Uniform LoRA matched and AdaLoRA in all five columns.
Second, at approximately half the rank-$32$ adapter budget, Adaptive exceeds the mean of Uniform LoRA at rank~$32$ on all five architectures and remains within the SEMs of Oracle Allocation, while staying $0.5$--$2.6$ points below Full Fine-Tuning across Tables~\ref{tab:allocation-consolidated} and~\ref{tab:allocation-calvin}.
We attribute the crossings to constrained-adaptation regularization.
Full Fine-Tuning and Uniform LoRA at rank~$32$ reach near-zero training loss within $3$ epochs while held-out validation loss rises from epoch $4$, whereas Adaptive's validation loss decreases monotonically.
Because we score every method at its best validation checkpoint (\S\ref{sec:exp-setup}), the crossings are not an artifact of overfit final checkpoints.
Even at its best checkpoint, Uniform LoRA at rank~$32$ trails Adaptive at twice the adapter budget, so the crossing reflects where capacity is placed rather than checkpoint choice.
Third, the advantage concentrates at low budgets (Fig.~\ref{fig:pareto-facets}), and we do not claim superiority at every budget.
Fourth, Random Allocation collapses at the same budget, so the gains come from where the diagnostic places capacity rather than from non-uniformity per se, and Surgical, which spends the whole budget on the dominant region, trails even Uniform LoRA matched under the combined shift, so graded assignment rather than dominant-region selection drives the gains.
The ordering transfers to CALVIN on all five architectures (Table~\ref{tab:allocation-calvin}).

\begin{table}[tbp]
    \centering
    \setlength{\abovecaptionskip}{0pt}
    \setlength{\belowcaptionskip}{2pt}
    \caption{Task success rate ($\%$) on CALVIN environment~D, mean $\pm$ SEM over $5$ seeds and $50$ rollouts per task. Budgets, bold, and $\dagger$/$\ddagger$ follow Table~\ref{tab:allocation-consolidated}.}
    \label{tab:allocation-calvin}
    \renewcommand{\arraystretch}{0.92}
    \resizebox{\columnwidth}{!}{%
    \begin{tabular}{@{}lccccc@{}}
        \toprule
        \textbf{Method} & \textbf{OpenVLA-OFT} (16M) & $\boldsymbol{\pi_0}$ (12M) & \textbf{SmolVLA} (3.4M) & \textbf{DTP} (2.6M) & \textbf{Octo} (1.4M) \\
        \midrule
        No Adaptation           & 19.7$\pm$3.8 & 22.4$\pm$2.9 & 15.8$\pm$3.3 & 18.9$\pm$4.4 & 14.6$\pm$3.6 \\
        Full Fine-Tuning        & 71.2$\pm$1.4 & 73.5$\pm$1.1 & 64.0$\pm$1.6 & 59.6$\pm$1.3 & 55.7$\pm$1.2 \\
        Uniform LoRA $r{=}32$   & 67.8$\pm$2.5 & 69.8$\pm$1.9 & 61.3$\pm$2.8 & 57.2$\pm$2.2 & 53.2$\pm$1.7 \\
        Oracle Allocation       & 70.4$\pm$1.2 & 71.6$\pm$0.9 & 62.5$\pm$1.3 & 58.5$\pm$1.0 & 54.6$\pm$1.1 \\
        \midrule
        Uniform LoRA matched    & 59.6$\pm$4.4 & 60.7$\pm$5.8 & 49.3$\pm$6.3 & 45.2$\pm$3.9 & 43.6$\pm$5.1 \\
        DoRA matched~\cite{Liu2024DoRA} & 61.9$\pm$2.6 & 64.2$\pm$4.1 & 51.0$\pm$3.3 & 48.9$\pm$5.0 & 45.4$\pm$4.6 \\
        $(IA)^3$~\cite{Liu2022FewShot}  & 46.2$\pm$7.4 & 50.9$\pm$5.2 & 38.1$\pm$6.8 & 39.7$\pm$5.9 & 34.3$\pm$9.6 \\
        Random Allocation       & 37.1$\pm$8.8 & 41.9$\pm$6.4 & 30.6$\pm$9.7 & 31.8$\pm$10.9 & 27.2$\pm$8.1 \\
        Surgical (dominant region) & 53.8$\pm$3.9 & 54.7$\pm$3.4 & 42.6$\pm$4.5 & 40.1$\pm$3.7 & 38.9$\pm$4.2 \\
        AdaLoRA~\cite{Zhang2023AdaLoRA} & 64.7$\pm$2.3 & 66.8$\pm$3.6 & 55.1$\pm$2.9 & 52.9$\pm$1.9 & 48.7$\pm$3.1 \\
        LoRA-SP~\cite{Kim2026LoRASP} & 64.5$\pm$2.7 & 68.6$\pm$2.0 & 56.4$\pm$4.4 & 55.8$\pm$2.2 & 42.9$\pm$2.4 \\
        \midrule
        \textbf{Adaptive} & \textbf{69.8$\pm$1.5}$^{\dagger\ddagger}$ & \textbf{72.1$\pm$1.2}$^{\dagger\ddagger}$ & \textbf{62.1$\pm$1.4}$^{\dagger\ddagger}$ & \textbf{58.3$\pm$1.0}$^{\dagger\ddagger}$ & \textbf{54.3$\pm$1.3}$^{\dagger\ddagger}$ \\
        \bottomrule
    \end{tabular}}
\end{table}

\newcommand{\pcell}[1]{\includegraphics[width=\linewidth]{images/pareto_cells/#1}}
\newcommand{\pW}{0.185\textwidth}
\newcommand{\parch}[2]{\begin{minipage}[c]{0.06\textwidth}\centering\textbf{\scriptsize #1}\\{\tiny (#2)}\end{minipage}}
\newcommand{\pshifthd}[1]{\begin{minipage}[c]{\pW}\centering\textbf{\footnotesize #1}\end{minipage}}
\newcommand{\pcellbox}[1]{\begin{minipage}[c]{\pW}\centering\pcell{#1}\end{minipage}}

\begin{figure*}[tp]
    \setlength{\abovecaptionskip}{2pt}
    \setlength{\belowcaptionskip}{-12pt}
    \centering
    \resizebox{0.83\textwidth}{!}{\begin{minipage}{\textwidth}
    \centering
    \setlength{\tabcolsep}{0.5pt}
    \renewcommand{\arraystretch}{0.85}
    \begin{tabular}{cccccc}
        & \pshifthd{Appearance} & \pshifthd{Spatial} & \pshifthd{Novel Object} & \pshifthd{Instruction} & \pshifthd{Combined} \\[-1pt]
        \parch{OpenVLA-OFT}{7B} &
        \pcellbox{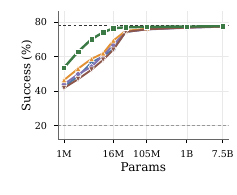} & \pcellbox{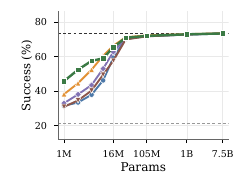} &
        \pcellbox{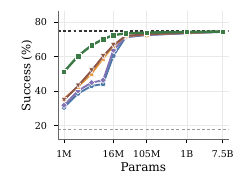} & \pcellbox{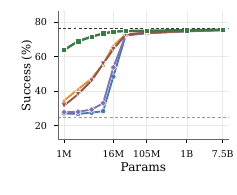} &
        \pcellbox{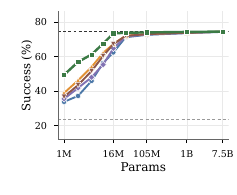} \\[-2pt]
        \parch{$\pi_0$}{3.3B} &
        \pcellbox{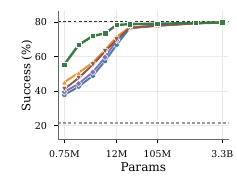} & \pcellbox{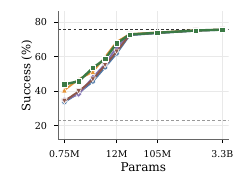} &
        \pcellbox{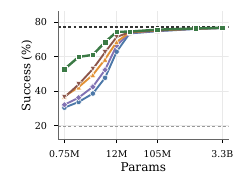} & \pcellbox{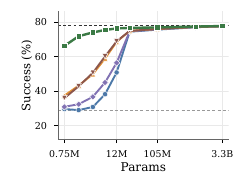} &
        \pcellbox{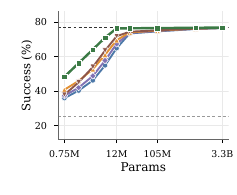} \\
    \end{tabular}\\[1pt]
    \includegraphics[width=0.9\textwidth]{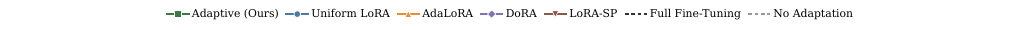}
    \end{minipage}}
    \caption{Parameter--performance frontiers on LIBERO for OpenVLA-OFT and $\pi_0$ under the five deployment shifts, sweeping Adaptive, Uniform LoRA, AdaLoRA, DoRA, and LoRA-SP over nine budgets from one sixteenth of the matched budget to the full model size; the remaining three architectures appear in the accompanying video. Dashed lines mark the No Adaptation floor and Full Fine-Tuning ceiling. The Adaptive advantage is largest at low budgets and narrows as capacity approaches rank~$32$.}
    \label{fig:pareto-facets}
\end{figure*}





\subsection{Ablation Study}
\label{sec:exp-ablations}

Table~\ref{tab:ablations-compact} ablates one design choice at a time.
On the allocation side, the freezing threshold $\delta = 0.1$ matches no freezing at the same budget and beats aggressive freezing; the value of freezing lies in budget rather than single-shift success, since frozen regions carry no adapter, which is what yields the $3.1$M language budget and the training-time parity with Uniform LoRA matched in Table~\ref{tab:physical-master}.
Replacing the diagnostic-driven assignment with equal or random ranks at the same budget costs $12$ and $31$ points of success, which supports the claim that placement, not budget, drives the gains.
Varying the gate temperature over two orders of magnitude, $\gamma \in [0.1, 10]$, changes success by less than one point, so the gate needs only coarse calibration.

\begin{table}[tbp]
    \centering
    \setlength{\abovecaptionskip}{0pt}
    \setlength{\belowcaptionskip}{-5pt}
    \caption{Ablations on OpenVLA-OFT, LIBERO combined shift, $16$M matched budget, $5$ seeds. Each row removes or alters one component of the full pipeline (Ours). $\rho_S$ and $R^2$ follow \S\ref{sec:exp-diagnostic}. Allocation-only rows leave the diagnostic unchanged and report success rate only. Equal allocation spreads the budget over the pipeline's region set and so differs in adapter placement from Uniform LoRA matched (Table~\ref{tab:allocation-consolidated}). Every ablated row differs significantly from the full pipeline at $p<0.05$ (paired t-test) except Ours $-$ freezing.}
    \label{tab:ablations-compact}
    \scriptsize
    \renewcommand{\arraystretch}{0.92}
    \setlength{\tabcolsep}{3.5pt}
    \begin{tabular}{@{}lccc@{}}
        \toprule
        \textbf{Variant} & $\boldsymbol{\rho_S}$ & $\boldsymbol{R^2}$ & \textbf{Success ($\%$)} \\
        \midrule
        \textbf{Ours (full pipeline)}             & \textbf{0.96$\pm$0.02} & \textbf{0.93$\pm$0.03} & \textbf{73.4$\pm$1.3} \\
        \midrule
        Ours $-$ Tier~2                           & 0.83$\pm$0.04 & 0.66$\pm$0.05 & 67.0$\pm$2.6 \\
        Ours $-$ Tier~1                           & 0.88$\pm$0.03 & 0.74$\pm$0.04 & 68.8$\pm$2.4 \\
        Ours $-$ gate (uniform $0.5$ blend)       & 0.92$\pm$0.03 & 0.85$\pm$0.04 & 70.8$\pm$1.7 \\
        Ours $-$ coreset (random cache)           & 0.85$\pm$0.05 & 0.70$\pm$0.06 & 69.2$\pm$2.9 \\
        Ours with $N = 1$                         & 0.71$\pm$0.09 & 0.47$\pm$0.10 & 64.2$\pm$4.7 \\
        \midrule
        Ours $-$ freezing ($\delta = 0$)          & --- & --- & 71.0$\pm$2.4 \\
        Ours with $\delta = 0.3$                  & --- & --- & 65.7$\pm$3.8 \\
        Ours $-$ diagnostic ranks (equal)         & --- & --- & 61.9$\pm$5.4 \\
        Ours $-$ diagnostic ranks (random)        & --- & --- & 42.8$\pm$7.7 \\
        \bottomrule
    \end{tabular}
\end{table}

\subsection{Physical Robot Evaluation}
\label{sec:exp-physical}

\begin{figure}[!tp]
    \centering
    \setlength{\abovecaptionskip}{4pt}
    \setlength{\belowcaptionskip}{-0pt}
    \includegraphics[width=0.95\columnwidth]{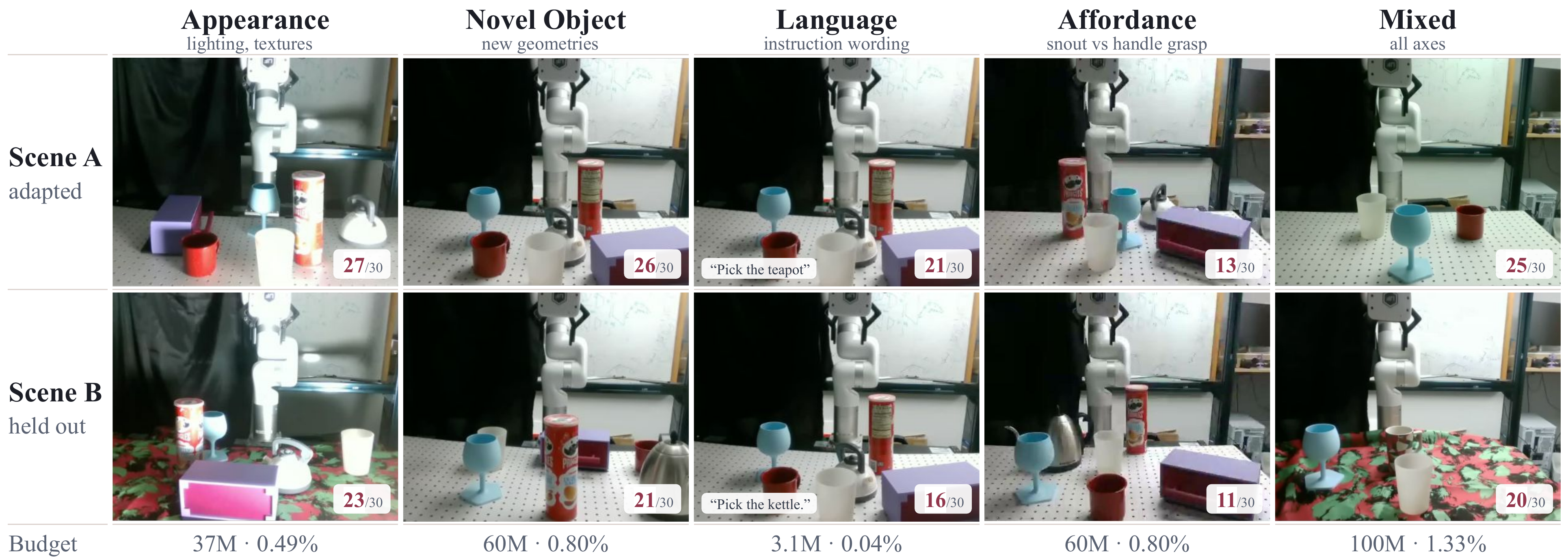}
    \caption{Five of the six xArm-7 shifts from the fixed third-person camera, with the adaptation condition (Scene~A, top) and the held-out condition (Scene~B, bottom) evaluated without retraining: appearance (lighting, then lighting plus background), novel object (toy and metallic teapot substitutions for the kettle), language (reworded instructions, with Scene~B naming the teapot), affordance (grasping the teapot by its snout rather than its handle), and mixed (a new cup geometry and color against a camouflaging background). Counts are Adaptive's successes of $30$ paired rollouts; the bottom row gives each shift's selected budget as a parameter count and a fraction of the $7.5$B policy.}
    \label{fig:physical-scenes}
\end{figure}

We evaluate the pipeline on a UFactory xArm-7 manipulator with a parallel-jaw gripper, OpenVLA-OFT as the base policy, one fixed third-person RGB camera viewing the workspace at a downward tilt, and one wrist-mounted RGB camera.
For each of six controlled shifts we collect $20$ examples under the shifted condition, fine-tune each method for $10$ epochs on one NVIDIA A100 80GB, and evaluate every method at the shared epoch-$10$ checkpoint on $30$ paired initial configurations held identical across methods; the shared checkpoint avoids per-method checkpoint selection on the small physical validation sets, and the pairing supports exact two-sided McNemar tests on discordant outcomes, reported per comparison without multiple-comparison correction.
For five of the six shifts we additionally evaluate a held-out Scene~B \emph{without} retraining.
Fig.~\ref{fig:physical-scenes} shows the shifts and Table~\ref{tab:physical-master} reports all results.
The diagnostic selects the budget per shift, spanning $32\times$ from $3.1$M for language to $100$M for mixed, and funds the DINOv2 and SigLIP encoders under appearance, SigLIP together with the action head under novel object and affordance, and rank-$3$ adapters on the late LLaMA region under language, whose per-rank cost sets the $3.1$M budget.
The diagnostic runs in roughly $90$ seconds on the same A100 for the $7$B model, against $20$ minutes of adapter training.

\begin{table}[tbp]
    \centering
    \setlength{\abovecaptionskip}{0pt}
    \setlength{\belowcaptionskip}{-9pt}
    \caption{Physical results on the xArm-7 with OpenVLA-OFT, successes out of $30$ paired rollouts at epoch~$10$ with identical initial configurations across methods (paired McNemar's tests). We evaluate Scene~B without retraining. \emph{Budget} is the per-shift budget the diagnostic selects, which the matched baselines and LoRA-SP share. Under the all-linear insertion, rank~$1$ costs ${\sim}3.4$M, so the matched baselines and LoRA-SP on the language shift run at ${\sim}3.4$M rather than $3.1$M. AdaLoRA runs at its own allocation ($45$--$105$M), Uniform LoRA $r{=}32$ at $110$M (the hardware recipe adapts every linear layer, not only attention projections), and Full Fine-Tuning updates all $7.5$B parameters. \emph{Wall} is training minutes for $10$ epochs on one A100 80GB, measured on the appearance shift and excluding the ${\sim}90$-second diagnostic. Bold marks the best parameter-efficient result per column; shading runs red (worst) to green (best) within each column.}
    \label{tab:physical-master}
    \renewcommand{\arraystretch}{1.05}
    \setlength{\tabcolsep}{2.0pt}
    \resizebox{\columnwidth}{!}{%
    \begin{tabular}{@{}lccccccccccccc@{}}
        \toprule
        & \multicolumn{2}{c}{\textbf{Appearance}} & \textbf{Spatial} & \multicolumn{2}{c}{\textbf{Novel Obj.}} & \multicolumn{2}{c}{\textbf{Language}} & \multicolumn{2}{c}{\textbf{Affordance}} & \multicolumn{2}{c}{\textbf{Mixed}} & \\
        \cmidrule(lr){2-3} \cmidrule(lr){4-4} \cmidrule(lr){5-6} \cmidrule(lr){7-8} \cmidrule(lr){9-10} \cmidrule(lr){11-12}
        \textbf{Method} & A & B & A & A & B & A & B & A & B & A & B & \textbf{Wall$~\downarrow$} \\
        \midrule
        Budget (selected)      & \multicolumn{2}{c}{$37$M} & $32$M & \multicolumn{2}{c}{$60$M} & \multicolumn{2}{c}{$3.1$M} & \multicolumn{2}{c}{$60$M} & \multicolumn{2}{c}{$100$M} & \\
        \midrule
        No Adaptation          & \cl{0}5/30   & \cl{0}1/30   & \cl{0}4/30   & \cl{0}5/30   & \cl{0}2/30   & \cl{0}5/30   & \cl{40}4/30  & \cl{0}0/30   & \cl{0}0/30   & \cl{0}7/30   & \cl{0}3/30   & ---    \\
        Full Fine-Tuning       & \ch{82}25/30 & \cl{72}9/30  & \ch{100}19/30 & \ch{80}24/30 & \cl{64}8/30  & \ch{100}22/30 & \ch{34}11/30 & \ch{70}11/30 & \cl{36}2/30  & \ch{88}24/30 & \cl{70}9/30  & \cl{0}55 \\
        Uniform LoRA $r{=}32$  & \ch{54}22/30 & \ch{36}16/30 & \ch{74}17/30 & \ch{42}20/30 & \ch{26}14/30 & \ch{52}18/30 & \ch{60}13/30 & \ch{38}9/30  & \cl{90}5/30  & \ch{66}22/30 & \ch{6}12/30  & \ch{88}22 \\
        Uniform LoRA matched   & \ch{36}20/30 & \ch{28}15/30 & \ch{46}15/30 & \ch{52}21/30 & \ch{36}15/30 & \cl{24}7/30  & \cl{0}1/30   & \ch{38}9/30  & \cl{90}5/30  & \ch{56}21/30 & \ch{6}12/30  & \ch{100}\textbf{20} \\
        DoRA matched           & \ch{46}21/30 & \ch{36}16/30 & \ch{74}17/30 & \ch{62}22/30 & \ch{48}16/30 & \cl{36}8/30  & \cl{26}3/30  & \ch{54}10/30 & \ch{10}6/30  & \ch{66}22/30 & \ch{18}13/30 & \ch{78}24 \\
        AdaLoRA                & \ch{54}22/30 & \ch{46}17/30 & \ch{100}\textbf{19/30} & \ch{62}22/30 & \ch{48}16/30 & \ch{64}19/30 & \ch{74}14/30 & \ch{54}10/30 & \ch{10}6/30 & \ch{78}23/30 & \ch{30}14/30 & \ch{72}25 \\
        LoRA-SP                & \ch{28}19/30 & \ch{54}18/30 & \cl{94}11/30 & \ch{62}22/30 & \ch{26}14/30 & \ch{18}15/30 & \ch{60}13/30 & \ch{38}9/30  & \ch{46}8/30  & \ch{34}19/30 & \ch{6}12/30  & \ch{66}26 \\
        \midrule
        \textbf{Adaptive}      & \ch{100}\textbf{27/30} & \ch{100}\textbf{23/30} & \ch{60}16/30 & \ch{100}\textbf{26/30} & \ch{100}\textbf{21/30} & \ch{88}\textbf{21/30} & \ch{100}\textbf{16/30} & \ch{100}\textbf{13/30} & \ch{100}\textbf{11/30} & \ch{100}\textbf{25/30} & \ch{100}\textbf{20/30} & \ch{100}\textbf{20} \\
        \bottomrule
    \end{tabular}}
\end{table}

The language shift is the sharpest case.
At the language budget, Uniform LoRA matched reaches $7/30$ against a No Adaptation floor of $5/30$, while Adaptive reaches $21/30$ against $22/30$ for Full Fine-Tuning at $0.04\%$ of its trainable parameters; AdaLoRA reaches $19/30$ using $45$M parameters, while LoRA-SP reaches $15/30$ at the matched language budget (Table~\ref{tab:physical-master}), so non-uniform placement is what survives the tight budget and pre-hoc diagnosis extends the margin.
Across the five held-out Scene-B conditions, Adaptive records $11$--$23$ successes of $30$, against $8$--$18$ for the strongest PEFT baseline in each condition and $2$--$11$ for Full Fine-Tuning, and it is numerically best among PEFT methods in all five Scene-B conditions and five of six Scene-A conditions.
Its margins over Full Fine-Tuning on the appearance, novel object, affordance, and mixed Scene-B conditions ($9$--$14$) are significant under McNemar's test, while its margins over the strongest PEFT baseline ($2$--$5$ on Scene~A, $2$--$6$ on Scene~B) are not individually significant at $n = 30$.
These results show favorable held-out performance under the shared epoch-$10$ protocol and do not establish superiority over separately early-stopped Full Fine-Tuning; what distinguishes Adaptive is delivering this at equal or smaller budgets ($3.1$--$100$M against AdaLoRA's $45$--$105$M).
On the affordance shift, No Adaptation returns $0/30$ on every column, so reported successes reflect part-grounded grasps, and Adaptive leads the best PEFT baseline by three successes on both scenes.
The physical spatial shift occludes the target itself, and its diagnostic profile spreads across mid-to-late vision rather than concentrating, unlike the simulated spatial shift, whose clutter leaves the target visible and loads trajectory adjustment in the action head; accordingly, no allocation strategy separates from the others, with Adaptive at $16/30$ against AdaLoRA's $19/30$ (not significant) while using $43\%$ of AdaLoRA's parameters.
We report this as the boundary of the claim, as the diagnostic helps when the importance signal concentrates and we detect no significant difference among allocation strategies otherwise.
\section{Conclusion}
\label{sec:conclusion}
This paper asked whether adaptation cost in VLA models is structured enough to measure before fine-tuning begins.
Across five architectures and five deployment shifts, the adaptation spectrum supports an affirmative answer, with each shift type concentrating cost in a characteristic functional region regardless of parameter scale or action-decoding paradigm in our experiments.
The diagnostic recovers this structure from ten unlabeled target observations without fine-tuning, ranking regions within each deployment at a median Spearman of $0.91$, and the allocator converts the prediction into variable-rank adapters that match or exceed uniform LoRA at every budget we tested.
On the physical xArm-7, the selected budgets span a $32\times$ range, and Adaptive achieves higher held-out-scene success than Full Fine-Tuning under the shared fixed-epoch protocol.

\textbf{Limitations and Future Work.}
The source-side Fisher can misestimate target sensitivity when the domain gap is large (the regime the gate routes to Tier~2), and the measured $\rho(r)$ depends on the fine-tuning protocol (\S\ref{sec:exp-spectrum}).
Hyperparameters calibrated once on LIBERO transferred unchanged to CALVIN and to the xArm-7, though we have not tested them beyond these settings, and the hardware evaluation covers one architecture and one embodiment.
The adapt stage still requires roughly $20$ labeled demonstrations, so ``unlabeled'' describes diagnosis, not adaptation.
The spatial-occlusion result marks where the pipeline stops helping, when the importance signal spreads rather than concentrates.


\bibliographystyle{IEEEtran}
\bibliography{references}

\begin{thebibliography}{10}
\providecommand{\url}[1]{#1}
\csname url@rmstyle\endcsname
\providecommand{\newblock}{\relax}
\providecommand{\bibinfo}[2]{#2}
\providecommand\BIBentrySTDinterwordspacing{\spaceskip=0pt\relax}
\providecommand\BIBentryALTinterwordstretchfactor{4}
\providecommand\BIBentryALTinterwordspacing{\spaceskip=\fontdimen2\font plus
\BIBentryALTinterwordstretchfactor\fontdimen3\font minus
  \fontdimen4\font\relax}
\providecommand\BIBforeignlanguage[2]{{%
\expandafter\ifx\csname l@#1\endcsname\relax
\typeout{** WARNING: IEEEtran.bst: No hyphenation pattern has been}%
\typeout{** loaded for the language `#1'. Using the pattern for}%
\typeout{** the default language instead.}%
\else
\language=\csname l@#1\endcsname
\fi
#2}}

\bibitem{Kim2024OpenVLA}
M.~J. Kim, \emph{et~al.}, ``{OpenVLA}: An open-source vision-language-action
  model,'' in \emph{Conference on Robot Learning (CoRL)}, 2024.

\bibitem{OctoModelTeam2024}
{Octo Model Team}, \emph{et~al.}, ``Octo: An open-source generalist robot
  policy,'' in \emph{Robotics: Science and Systems (RSS)}, 2024.

\bibitem{Ferchau2026VLALoRA}
F.~Ferchau, D.~Pommer, and C.~Axenie, ``On the efficiency of {LoRA} fine-tuning
  for vision-language-action models in industrial robotic manipulation,''
  \emph{arXiv:2607.10172}, 2026.

\bibitem{Hu2022LoRA}
E.~J. Hu, \emph{et~al.}, ``{LoRA}: Low-rank adaptation of large language
  models,'' in \emph{International Conference on Learning Representations
  (ICLR)}, 2022.

\bibitem{Wolpert1998MOSAIC}
D.~M. Wolpert and M.~Kawato, ``Multiple paired forward and inverse models for
  motor control,'' \emph{Neural Networks}, vol.~11, no. 7--8, pp. 1317--1329,
  1998.

\bibitem{Lee2023Surgical}
Y.~Lee, \emph{et~al.}, ``Surgical fine-tuning improves adaptation to
  distribution shifts,'' in \emph{International Conference on Learning
  Representations (ICLR)}, 2023.

\bibitem{He2023SPT}
H.~He, J.~Cai, J.~Zhang, D.~Tao, and B.~Zhuang, ``Sensitivity-aware visual
  parameter-efficient fine-tuning,'' in \emph{IEEE/CVF International Conference
  on Computer Vision (ICCV)}, 2023.

\bibitem{Zhang2023AdaLoRA}
Q.~Zhang, \emph{et~al.}, ``{AdaLoRA}: Adaptive budget allocation for
  parameter-efficient fine-tuning,'' in \emph{International Conference on
  Learning Representations (ICLR)}, 2023.

\bibitem{Kim2026LoRASP}
D.~Kim, \emph{et~al.}, ``Adaptive capacity allocation for vision language
  action fine-tuning,'' \emph{arXiv:2603.07404}, 2026.

\bibitem{Shi2026VLATrace}
H.~Shi, X.~Ren, Y.~Zhang, \emph{et~al.}, ``{VLA-Trace}: Diagnosing
  vision-language-action models through representation and behavior tracing,''
  \emph{arXiv:2605.30117}, 2026.

\bibitem{Liu2023LIBERO}
B.~Liu, \emph{et~al.}, ``{LIBERO}: Benchmarking knowledge transfer for lifelong
  robot learning,'' in \emph{Adv. Neural Inf. Process. Syst. (NeurIPS)}, 2023.

\bibitem{Mees2022CALVIN}
O.~Mees, L.~Hermann, E.~Rosete-Beas, and W.~Burgard, ``{CALVIN}: A benchmark
  for language-conditioned policy learning for long-horizon robot manipulation
  tasks,'' \emph{IEEE Robot. Autom. Lett.}, vol.~7, 2022.

\bibitem{Brohan2023RT1}
A.~Brohan, \emph{et~al.}, ``{RT-1}: Robotics transformer for real-world control
  at scale,'' in \emph{Robotics: Science and Systems (RSS)}, 2023.

\bibitem{Kim2025OpenVLAOFT}
M.~J. Kim, C.~Finn, and P.~Liang, ``Fine-tuning vision-language-action models:
  Optimizing speed and success,'' \emph{arXiv:2502.19645}, 2025.

\bibitem{Hou2024DTP}
Z.~Hou, \emph{et~al.}, ``Diffusion transformer policy,''
  \emph{arXiv:2410.15959}, 2024.

\bibitem{Black2024Pi0}
K.~Black, \emph{et~al.}, ``$\pi_0$: A vision-language-action flow model for
  general robot control,'' \emph{arXiv:2410.24164}, 2024.

\bibitem{Shukor2025SmolVLA}
M.~Shukor, \emph{et~al.}, ``{SmolVLA}: A vision-language-action model for
  affordable and efficient robotics,'' \emph{arXiv:2506.01844}, 2025.

\bibitem{Aghajanyan2021Intrinsic}
A.~Aghajanyan, L.~Zettlemoyer, and S.~Gupta, ``Intrinsic dimensionality
  explains the effectiveness of language model fine-tuning,''
  \emph{arXiv:2012.13255}, 2021.

\bibitem{Biderman2024LoRALearnsLess}
D.~Biderman, \emph{et~al.}, ``{LoRA} learns less and forgets less,''
  \emph{arXiv:2405.09673}, 2024.

\bibitem{ARDLoRA2025}
H.~U.~K. Shinwari and M.~Usama, ``{ARD-LoRA}: Dynamic rank allocation for
  parameter-efficient fine-tuning of foundation models with heterogeneous
  adaptation needs,'' \emph{arXiv:2506.18267}, 2025.

\bibitem{Liu2026RSRA}
J.~Liu, H.~Kang, Q.~Zhao, G.~Yu, and J.~Wang, ``{RSRA}: Training-free probing
  of representation sensitivity for efficient {LoRA} rank allocation,''
  \emph{arXiv:2607.09757}, 2026.

\bibitem{Kornblith2019CKA}
S.~Kornblith, M.~Norouzi, H.~Lee, and G.~Hinton, ``Similarity of neural network
  representations revisited,'' in \emph{International Conference on Machine
  Learning (ICML)}, 2019.

\bibitem{Dowson1982Frechet}
D.~C. Dowson and B.~V. Landau, ``The {Fr\'{e}chet} distance between
  multivariate normal distributions,'' \emph{J. Multivariate Anal.}, vol.~12,
  1982.

\bibitem{Gal2016MCDropout}
Y.~Gal and Z.~Ghahramani, ``Dropout as a {Bayesian} approximation: Representing
  model uncertainty in deep learning,'' in \emph{International Conference on
  Machine Learning (ICML)}, 2016.

\bibitem{Martens2015KFAC}
J.~Martens and R.~Grosse, ``Optimizing neural networks with
  {Kronecker}-factored approximate curvature,'' in \emph{International
  Conference on Machine Learning (ICML)}, 2015.

\bibitem{HarPeled2007Coresets}
S.~Har-Peled and A.~Kushal, ``Smaller coresets for $k$-median and $k$-means
  clustering,'' \emph{Discrete Comput. Geom.}, vol.~37, 2007.

\bibitem{Liu2024DoRA}
S.-Y. Liu, \emph{et~al.}, ``{DoRA}: Weight-decomposed low-rank adaptation,''
  \emph{arXiv:2402.09353}, 2024.

\bibitem{Liu2022FewShot}
H.~Liu, \emph{et~al.}, ``Few-shot parameter-efficient fine-tuning is better and
  cheaper than in-context learning,'' in \emph{Adv. Neural Inf. Process. Syst.
  (NeurIPS)}, 2022.

\end{thebibliography}

\end{document}